\documentclass[letterpaper]{article} % DO NOT CHANGE THIS
\usepackage{aaai25}  % DO NOT CHANGE THIS
\usepackage{times}  % DO NOT CHANGE THIS
\usepackage{helvet}  % DO NOT CHANGE THIS
\usepackage{courier}  % DO NOT CHANGE THIS
\usepackage[hyphens]{url}  % DO NOT CHANGE THIS
\usepackage{graphicx} % DO NOT CHANGE THIS
\usepackage{natbib}  % DO NOT CHANGE THIS AND DO NOT ADD ANY OPTIONS TO IT
\usepackage{caption} % DO NOT CHANGE THIS AND DO NOT ADD ANY OPTIONS TO IT
\usepackage{algorithm}
\usepackage{algorithmic}
\usepackage{amsmath, amssymb, amsfonts}
\usepackage{tikz}
\usetikzlibrary{arrows.meta,positioning,shapes.geometric,fit,backgrounds}
\usepackage{booktabs}
\usepackage{multirow}
\usepackage{newfloat}
\usepackage{listings}
\DeclareCaptionStyle{ruled}{labelfont=normalfont,labelsep=colon,strut=off} % DO NOT CHANGE THIS
\floatstyle{ruled}
\newfloat{listing}{tb}{lst}{}
\floatname{listing}{Listing}
\title{MESD: A Risk-Sensitive Metric for Explanation Fairness Across Intersectional Subgroups}
\author{
    Gideon Popoola$^{1}$ and John Sheppard$^{1}$
}
\affiliations{
    \textsuperscript{\rm 1}Gianforte School of Computing
    Montana State University\\
    Bozeman, MT, USA, 59715\\
    gideon.popoola@student.montana.edu, john.sheppard@montana.edu
}

\usepackage{bibentry}
\begin{document}

\maketitle

\begin{abstract}
Fairness in machine learning is predominantly evaluated through outcome-oriented metrics, such as Demographic parity, which measure whether predictions are statistically consistent across protected groups. However, these metrics cannot detect whether a model uses systematically different reasoning for different demographic groups, which violates procedural fairness principles. This problem is compounded by intersectionality, where models may appear fair on individual attributes (e.g., race) while exhibiting significant disparities for intersectional subgroups (e.g., race $\times$ gender), a phenomenon known as fairness gerrymandering. In this work, we introduce Multi-category Explanation Stability Disparity (MESD), a procedural fairness metric that quantifies disparities in explanation quality across intersectional subgroups formed by the Cartesian product of multiple protected attributes. MESD integrates three components, which are label-aware aggregation aligned with outcome-conditional fairness, empirical-Bayes shrinkage to stabilize estimates for small intersectional groups, and Conditional Value-at-Risk (CVaR) weighting to emphasize worst-case subgroup disparities. We integrate MESD within a multi-objective optimization framework (UEF) that jointly optimizes utility, outcome fairness, and procedural fairness using NSGA-II. We evaluated MESD and UEF on three benchmark datasets along with four state-of-the-art methods in several experiments, and we demonstrate that MESD reveals procedural disparities invisible to outcome metrics alone. We position our contribution within procedural justice theory and discuss implications for regulatory compliance and intersectional equity.
\end{abstract}

% Uncomment the following to link to your code, datasets, an extended version or similar.
%
% \begin{links}
%     \link{Code}{https://aaai.org/example/code}
%     \link{Datasets}{https://aaai.org/example/datasets}
%     \link{Extended version}{https://aaai.org/example/extended-version}
% \end{links}

\section{Introduction}
Machine learning (ML) systems are increasingly used in socially sensitive domains such as credit decision-making, recidivism prediction, and health care resource allocation, which affect people's lives \cite{jayatilake2021involvement, barocas2016big, wang2023pursuit}. As these systems grow larger, concerns about fairness have produced a rich body of work on outcome-oriented metrics that evaluate whether predictions satisfy statistical constraints across protected groups \cite{pessach2022review, loukas2023demographic}. Metrics such as Demographic Parity \cite{hertweck2021moral} and Equalized Odds \cite{hardt2016equality} have become standard tools for auditing algorithmic bias.
 
However, outcome-oriented fairness can only reveal disparities in the model's outcomes, not how the model generates them. Consider two credit applicants, where one is a White male and the other a Black female, who both receive loan approval from the same model (Figure \ref{fig:motivation}). Outcome-oriented metrics (e.g., demographic parity) show that the model decision is fair because both received the favorable outcome. However, when we examine the model's explanations using post-hoc methods such as SHAP \cite{lundberg2017unified} or LIME \cite{ribeiro2016should}, the reasoning diverges sharply. For the first applicant, the model emphasizes income and credit history, which are genuine credit-worthiness features with clear relevance. For the second, the model relies on zip code and debt ratio, which are proxies that may encode historical disadvantage. The illustration shows that the outcome is equal, but the procedure is biased. 

\begin{figure}[t]
\centering
\begin{tikzpicture}[
    scale=0.82,
    transform shape,
    font=\scriptsize,
    card/.style={draw, rounded corners=5pt, thick, align=left, minimum width=3.7cm, minimum height=2.7cm},
    model/.style={draw, rounded corners=4pt, thick, fill=gray!20, align=center, minimum width=5.4cm, minimum height=0.7cm},
    outbox/.style={draw, rounded corners=4pt, thick, fill=green!50!black, text=white, align=center, minimum width=1.6cm, minimum height=0.55cm},
    summary/.style={draw, rounded corners=5pt, thick, fill=gray!12, align=center, minimum width=7.7cm, minimum height=1.0cm},
    arrow/.style={->, thick},
    pos/.style={text=green!50!black},
    neg/.style={text=red!70!black}
]

% title
\node[font=\bfseries] at (0,5.2) {Same decision, different explanation};

% applicant cards
\node[card, fill=teal!18] (A) at (-2.15,3.6) {};
\node[font=\bfseries] at (-2.15,4.45) {Applicant A};
\node[text=teal!60!black] at (-2.15,4.0) {White, Male, Age 35};

\node[anchor=west] at (-3.8,3.55) {Income:};
\node[anchor=east, font=\bfseries] at (-0.5,3.55) {\$72k};

\node[anchor=west] at (-3.8,3.15) {Credit hist.:};
\node[anchor=east, font=\bfseries] at (-0.5,3.15) {12 yrs};

\node[anchor=west] at (-3.8,2.75) {DTI:};
\node[anchor=east, font=\bfseries] at (-0.5,2.75) {0.28};

\node[anchor=west] at (-3.8,2.35) {Education:};
\node[anchor=east, font=\bfseries] at (-0.5,2.35) {Bachelor's};

\node[card, fill=orange!18] (B) at (2.15,3.6) {};
\node[font=\bfseries] at (2.15,4.45) {Applicant B};
\node[text=orange!70!black] at (2.15,4.0) {Black, Female, Age 34};

\node[anchor=west] at (0.5,3.55) {Income:};
\node[anchor=east, font=\bfseries] at (3.8,3.55) {\$72k};

\node[anchor=west] at (0.5,3.15) {Credit hist.:};
\node[anchor=east, font=\bfseries] at (3.8,3.15) {12 yrs};

\node[anchor=west] at (0.5,2.75) {DTI:};
\node[anchor=east, font=\bfseries] at (3.8,2.75) {0.28};

\node[anchor=west] at (0.5,2.35) {Education:};
\node[anchor=east, font=\bfseries] at (3.8,2.35) {Bachelor's};

% model
\node[model] (M) at (0,1.2) {Credit scoring model (black box)};
\draw[arrow] (A.south) -- (-2.15,1.55);
\draw[arrow] (B.south) -- (2.15,1.55);

% outcomes
\node[outbox] (OA) at (-2.15,0.0) {Approved};
\node[outbox] (OB) at (2.15,0.0) {Approved};

\draw[arrow] (M.south west) -- (-2.15,0.35);
\draw[arrow] (M.south east) -- (2.15,0.35);

\node[font=\bfseries] at (0,0.0) {$=$};

% explanation cards
\node[card, fill=teal!18] (EA) at (-2.15,-2.6) {};
\node[font=\bfseries] at (-2.15,-1.75) {Explanation A};
\node[anchor=west] at (-3.8,-2.15) {Income};
\node[anchor=east, font=\bfseries] at (-0.5,-2.15) {+0.42};
\node[anchor=west] at (-3.8,-2.55) {Credit hist.};
\node[anchor=east, font=\bfseries] at (-0.5,-2.55) {+0.35};
\node[anchor=west] at (-3.8,-2.95) {DTI};
\node[anchor=east, font=\bfseries] at (-0.5,-2.95) {-0.12};
\node[anchor=west] at (-3.8,-3.35) {Zip code};
\node[anchor=east, font=\bfseries] at (-0.5,-3.35) {+0.03};

\node[card, fill=orange!18] (EB) at (2.15,-2.6) {};
\node[font=\bfseries] at (2.15,-1.75) {Explanation B};
\node[anchor=west] at (0.5,-2.15) {Income};
\node[anchor=east, font=\bfseries] at (3.8,-2.15) {+0.15};
\node[anchor=west] at (0.5,-2.55) {Credit hist.};
\node[anchor=east, font=\bfseries] at (3.8,-2.55) {+0.11};
\node[anchor=west] at (0.5,-2.95) {DTI};
\node[anchor=east, font=\bfseries, neg] at (3.8,-2.95) {-0.31};
\node[anchor=west] at (0.5,-3.35) {Zip code};
\node[anchor=east, font=\bfseries] at (3.8,-3.35) {+0.28};

\draw[arrow] (OA.south) -- (-2.15,-1.25);
\draw[arrow] (OB.south) -- (2.15,-1.25);

\node[font=\bfseries] at (0,-2.75) {$\neq$};

% summary
\node[summary] at (0,-4.7) {\textbf{Procedural unfairness:} same outcome, but the model relies on different reasoning across groups.};

\end{tikzpicture}
\caption{Motivating example of procedural unfairness. Two applicants with nearly identical non-protected financial profiles receive the same decision (approved), yet the model assigns substantially different reasoning. The model uses income and credit history for Applicant A, but relies on zip code and debt ratio for Applicant B. Outcome-oriented fairness metrics detect no disparity, while our metric is designed to quantify such explanation disparities across protected and intersectional groups.}
\label{fig:motivation}
\end{figure}

This scenario illustrates a growing recognition that \emph{procedural fairness}, which deals with the fidelity, stability, and consistency of a model's reasoning, is a distinct and important dimension of algorithmic accountability \cite{grabowicz2022marrying, dai2022fairness, germino2025explanation}. A model satisfying Equalized Odds while using unstable or group-dependent reasoning violates principles of procedural justice foundational to legal and ethical frameworks \cite{shulner2022fairness, rawls1971egalitarian, thibaut1973procedural}.
 
The challenge is compounded by \emph{intersectionality}. Most fairness work considers protected attributes in isolation \cite{barocas2017fairness}, but individuals occupy multiple social categories simultaneously. A model may appear fair on race and gender separately yet exhibit significant disparities for intersectional subgroups such as Black women, a phenomenon Kearns et al.~\cite{kearns2018preventing} formalized as \emph{fairness gerrymandering}. This mirrors longstanding concerns in legal scholarship that single-axis analysis obscures the experiences of multiply-marginalized individuals~\cite{crenshaw1989demarginalizing}.

This paper addresses these gaps by developing the Multi-category Explanation Stability Disparity (MESD), a procedural fairness metric designed to operate across intersectional groups using multiple post-hoc explainers and label-aware aggregation. MESD assesses whether individuals across different intersectional groups receive explanations of comparable stability, consistency, and robustness when subjected to perturbations. To the best of our knowledge, MESD is among the first metrics to explicitly combine intersectional subgroup analysis with explanation stability and risk-sensitive aggregation. Also, we integrate MESD into a multi-objective optimization framework that optimizes utility, fairness, and explanation (UEF). This optimization framework reveals trade-offs that are invisible under outcome-oriented fairness alone and provides models that are simultaneously accurate, statistically fair, and procedurally fair. The contributions of this work are summarized as follows:
\begin{enumerate}
\item \textbf{MESD}: a procedural fairness metric for intersectional subgroups that incorporates label-aware aggregation, empirical-Bayes shrinkage, and CVaR-weighted tail-risk measurement to measure explanation fairness and prevent fairness gerrymandering.
\item \textbf{UEF}: a multi-objective optimization framework jointly optimizing utility, outcome fairness, and procedural fairness via NSGA-II with Chebyshev scalarization to produce a holistic model that is fair in both outcome-oriented and procedural fairness.
\item \textbf{Comprehensive validation}: Our results demonstrate that considering MESD can lead to a more equitable model. Also, UEF consistently achieves competitive or superior trade-offs across Pareto fronts (w.r.t. balancing outcome-oriented, procedural fairness, and utility) compared to established baselines. %59\% of the time, %nearly 3.5x better than the second-best method. 
\end{enumerate}

\section{Related Work}
\subsection{Outcome-Oriented Fairness}
Group fairness research has produced a taxonomy of metrics measuring whether predictions satisfy statistical properties across demographic groups. Demographic Parity \cite{dwork2012fairness} requires positive prediction rates to be independent of group membership, while Equalized Odds \cite{hardt2016equality} requires equal true-positive and false-positive rates. Algorithmic interventions include re-weighting training samples \cite{kamiran2012data}, adversarial debiasing \cite{zhang2018mitigating}, constrained optimization via reductions \cite{agarwal2018reductions}, and post-processing calibration \cite{hardt2016equality}. However, these approaches evaluate only \emph{what} the model predicts, not \emph{how} it arrives at predictions. As Zhao et al. \cite{zhao2023fairness} demonstrate, a model achieving outcome-oriented fairness may still evaluate protected groups using systematically different criteria, creating incentive structures that perpetuate societal biases.
 
\subsection{Intersectional Fairness}
 
Single-attribute fairness analysis can mask disparities affecting multiply-marginalized individuals. Crenshaw \cite{crenshaw1989demarginalizing} first articulated how single-axis legal frameworks fail to capture the unique discrimination experienced at the intersection of race and gender. In ML, Kearns et al. \cite{kearns2018preventing} formalized fairness gerrymandering, demonstrating that models can appear fair on each attribute individually while remaining unfair for intersections. Subsequent computational approaches include distributionally robust optimization \cite{himmelreich2024intersectionality, hertweck2024s}, probabilistic overlapping-group fairness \cite{yang2020fairness}, and resampling strategies \cite{popoola2024investigating}. Our work extends intersectional fairness from outcome-oriented fairness to procedural fairness, which is a new dimension to our knowledge, and no prior work addresses.
 
\subsection{Procedural Fairness and Explanation Quality}
The connection between explainability and fairness has received growing attention. Dai et al. \cite{dai2022fairness} introduced evaluation frameworks for evaluating disparities in explanation quality, demonstrating that models satisfying fairness criteria can exhibit qualitatively different reasoning across groups. Zhao et al. \cite{zhao2023fairness} proposed bridging fairness and explainability through explanation-aware model selection. Balagopalan et al. \cite{balagopalan2022road} showed that explanations themselves can reflect social biases. Grgic-Hlaca et al. \cite{grgic2018beyond} used feature selection to detect and mitigate procedural unfairness, relying on human feedback to determine which features are appropriate for decision-making. Wang, Huang, and Yao~\cite{wang2024procedural} proposed group procedural fairness as a measure using feature attribute explanations by measuring distances between explanations for similar members of different groups. Wang and Wu \cite{wang2024equalized} proposed equalized explainability, measuring explanation differences without considering counterfactuals or traditional fairness in optimization.
Most recently, Germino et al.~\cite{germino2025explanation} proposed Explanation Difference (EDiff), which measures the absolute difference in feature importance between counterfactual samples, and demonstrated the importance of treating fairness as a multi-objective optimization problem. EDiff represents a significant advance in procedural fairness measurement, but is limited to single binary protected attributes and counterfactual-based comparison.

\subsection{Explanation Stability and Ensemble Methods}
 
Post-hoc explanation methods such as Shapley Additive Explanation (SHAP) \cite{lundberg2017unified} and Local Interpretable Model-Agnostic Explanation (LIME) \cite{ribeiro2016should} are known to exhibit instability. Small perturbations to an input can produce substantially different explanations \cite{slack2020fooling}. To mitigate the instability in post-hoc methods, Bhatt et al. \cite{bhatt2020evaluating} proposed ensembling multiple explainers to improve consistency and reduce adversarial vulnerability. Begley et al. \cite{begley2020explainability} proposed stability metrics for individual explanations. 
MESD builds on these stability definitions while introducing intersectional group-level aggregation and distributional robustness, combining ideas from the stability and fairness literature that have historically evolved independently.
 
\subsection{Multi-Objective Fairness Optimization}
Fairness optimization is increasingly framed as a multi-objective problem. Wei and Niethammer~\cite{wei2022fairness} characterized the fairness-accuracy Pareto front. Wang et al.~\cite{wang2024generating} proposed generating diagnostic explanations for fair graph neural networks using multi-objective frameworks using evolutionary algorithms. Germino et al.~\cite{germino2025explanation} demonstrated a three-part loss function optimizing predictive performance, statistical parity, and explanation difference simultaneously. However, prior multi-objective approaches typically optimize only two objectives (utility vs.\ fairness) or consider only single protected attributes. Our UEF framework extends this to three objectives across intersectional groups.

Overall, our proposed method differs from prior work in three key respects. First, it operates on \emph{intersectional subgroups} formed by the Cartesian product of multiple protected attributes, detecting fairness gerrymandering at the intersection. Second, it measures \emph{stability under perturbation} rather than counterfactual difference, capturing whether explanations are robust and consistent within groups. Third, it incorporates distributional robustness through CVaR-weighted aggregation, ensuring that the metric does not mask extreme procedural disparities in small subgroups, which is a common phenomenon in multiple protected attributes.
 
\subsection{Ethical Positioning of Our Proposed MESD}
Before presenting the technical details of MESD, we ground our work in the procedural justice literature to clarify \emph{why} measuring explanation disparity constitutes a fairness concern.
 
The procedural justice tradition, originating with Thibaut and Walker \cite{thibaut1975procedural}, distinguishes between two fundamental types of justice: \emph{distributive justice} (fairness of outcomes) and \emph{procedural justice} (fairness of the processes used to reach outcomes). Their research demonstrated that people are willing to accept unfavorable outcomes when they perceive the process as fair. Rawls \cite{rawls1971theory, rawls1971egalitarian} further argued that justice requires not only fair outcomes but procedures that treat similar individuals similarly. This phenomenon was termed \emph{pure procedural justice}, in which the fairness of an outcome is entirely determined by the fairness of the process.
 
Translating these principles to ML, given two individuals $x_1$ and $x_2$ who differ only in protected attributes, a procedurally fair model should not only produce similar outcomes $f(x_1) \approx f(x_2)$, but these outcomes should arise from similar reasoning processes. Formally, the explanation functions should satisfy $|\Phi(x_1) - \Phi(x_2)| \approx 0$, where $\Phi$ represents the explanation function.
 
MESD implements this principle for intersectional groups by measuring whether the \emph{stability} of explanations is consistent across intersectional subgroups (e.g Race $\times$ Gender). The choice of stability (rather than direct explanation comparison) is deliberate because explanations for different individuals \emph{should} differ because individuals have different features. What should \emph{not} differ is the \emph{stability} and \emph{consistency} of those explanations. A model that provides stable, robust explanations for White males but noisy, inconsistent explanations for Black females violates procedural fairness even if its predictions are statistically fair. Also, following Crenshaw \cite{crenshaw1989demarginalizing}, we argue that procedural fairness must be evaluated at the intersection of protected attributes, not on each protected attribute independently. A model may provide stable explanations for women as a group and for Black individuals as a group, while providing systematically unstable explanations for Black women specifically. 
 
We acknowledge, following Binns \cite{binns2018fairness} and Green \cite{green2020false}, that technical fairness metrics, such as MESD, cannot fully resolve deeply rooted social inequalities. We position MESD not as a complete solution to procedural discrimination, but as one component in a holistic approach to ethical AI that must include participatory design, appropriate governance, and ongoing human oversight.

\section{The MESD Method}\label{meth}
We propose Multi-category Explanation Stability Disparity (MESD) as a procedural fairness metric that detects and mitigates procedural bias across multiple protected attributes. MESD builds on earlier work on the definitions of explanation stability, robustness, and sensitivity \cite{begley2020explainability}, while also incorporating novel concepts such as intersectional subgroups, label-aware aggregation, empirical-Bayes shrinkage, and distributionally robust CVaR-based measurement of worst-case disparities. MESD steps are explained below and in Algorithm \ref{alg:mesd}.

\subsection{Intersectional Subgroups}\label{cartessian} 
In the real world, humans can belong to more than one protected category (e.g, race and gender) \cite{kearns2018preventing}. To capture this more detailed bias in our metric and to prevent fairness gerrymandering, MESD constructs intersection protected groups via the Cartesian product of protected features. Let $\mathcal{A}_1, \mathcal{A}_2, \dots, \mathcal{A}_m$ represent $m$ protected features. Let
%MESD defines the set of intersectional groups as the Cartesian product:
\[
\mathcal{G} = \mathcal{A}_1 \times \mathcal{A}_2 \times \cdots \times \mathcal{A}_m.
\]
Each element $g \in \mathcal{G}$ corresponds to a unique intersectional subgroup (e.g., Black woman), and every individual is mapped to exactly one such subgroup via the protected attributes vector:
\begin{equation}
g(x) = (A_1(x), A_2(x), \ldots, A_m(x))
\end{equation}
However, such decomposition introduces statistical challenges. As $(|\mathcal{G}|)$ grows with the number of protected features, many intersectional subgroups may contain few samples, producing high-variance estimates of explanation stability. 

\subsection{Aggregation of Explainers}\label{agg}
Let $\phi^{(e)}(x) \in \mathbb{R}^d$ denote the normalized attribution vector returned by explainer $(e \in \mathcal{E})$, where $(\mathcal{E})$ is the set of chosen explainers (e.g., SHAP, LIME). Each attribution vector is normalized using signed $(L_1)$-normalization, ensuring comparability across explainers despite differing scales or sparsity patterns.
%\[
%\tilde{\phi}^{(e)}(x) = \frac{\phi^{(e)}(x)}{||\phi^{(e)}(x)||_1 + \varepsilon}
%\]
We then define an ensemble attribution vector as a convex combination:
\begin{equation}
\phi_{\mathrm{ens}}(x) = \sum_{e \in \mathcal{E}} w_e {\phi}^{(e)}(x),
\label{eq:phi_ens}
\end{equation}
where $(w_e \ge 0)$ and $(\sum_{e} w_e = 1)$. The weights $(w_e)$ can be user-specified 
%(e.g., equal weighting) 
or optimized to reflect the reliability of each explainer. 
This serves as a stabilizing mechanism to address inconsistencies that may arise from an explainer \cite{slack2020fooling}. 
Prior work has shown that explanation aggregation can improve consistency and reduce adversarial vulnerability \cite{bhatt2020evaluating}.

\subsection{Perturbation}
To evaluate the robustness, stability, and consistency of an explanation, we generate several perturbed samples of the inputs to test whether explanations generated by a model remain stable under small perturbations. 
Given an instance $x \in \mathbb{R}^d$, we generate a set of perturbed variants $x^{(1)}, \ldots, x^{(K)}$. Perturbations should be sufficiently small so that $x^{(k)}$ remains within a local neighborhood of $x$. 
We adopt a hybrid perturbation framework that combines Gaussian noise and stochastic feature masking. For Gaussian perturbations, we sample
\[
x^{(k)} = x + \eta^{(k)}, \qquad \eta^{(k)} \sim \mathcal{N}(0, \sigma^2 I_d)
\]
%This formulation captures the idea that explanations should not fluctuate significantly when minor variations occur in the input. 
The masking perturbation replaces randomly selected features with a baseline value $b_j$:
\[
x^{(k)}_j =
\begin{cases}
x_j, \text{with probability } 1 - p_m, \\
b_j, \text{with probability } p_m,
\end{cases}
\]
where $p_m$ is the masking probability and $b_j$ is a baseline value (e.g., the feature mean). Masking simulates missingness or partial erasure and tests whether explanations remain consistent when certain nonessential inputs are missing.

%For each perturbation generated, we compute the corresponding normalized attribution vector $(\tilde{\phi}(x^{(k)}))$. Explanation stability is quantified by comparing these perturbed explanations to the original:
%\begin{equation}
%\kappa(x) = \frac{2}{K(K+1)} \sum_{1 \le r < k \le K+1}
%\mathrm{Euc}\bigl(\tilde{\phi}^{(r)}, \tilde{\phi}^{(k)}\bigr),
%\end{equation}
%where $(\tilde{\phi}^{(1)} = \tilde{\phi}(x))$ is the base explanation and the remaining vectors correspond to perturbed inputs. Euclidean distance is used to measure stability because it is intuitive and straightforward. A high value of $\kappa(x)$ indicates high stability in the model explanation, whereas a low value of $\kappa(x)$ signals low stability or inconsistent reasoning.

\subsection{Stability Evaluation}
To quantify the stability of each instance, we used Euclidean distance to measure the distance between the explanations of the original and perturbed instances,
%. The rationale for this stability is that if the explanation is stable, then 
under the assumption that 
the distance between the original and perturbed explanation should be insignificant.

Formally, consider an instance $x \in \mathbb{R}^d$ with ensemble explanation $(\phi_{\mathrm{ens}}(x))$ defined in Equation \ref{eq:phi_ens}. We construct a local perturbation neighborhood $\mathcal{N}(x)$ consisting of (K) perturbed variants of $x$, denoted ${x^{(1)}, \dots, x^{(K)}}$. Perturbations are generated with Gaussian perturbations and feature masking:
\[
x^{(k)} = x + \eta^{(k)} - m^{(k)} \odot x + m^{(k)} \odot b
\]
%where $(\eta^{(k)} \sim \mathcal{N}(0, \sigma^2 I))$ injects small continuous variation, $(m^{(k)})$ is a masking vector with entries sampled independently from $(\mathrm{Bernoulli}(p))$, $(b)$ is a baseline vector (typically zeros or dataset means), $(\odot)$ denotes element-wise multiplication.
For each perturbed instance, a corresponding ensemble explanation is computed as $\phi_{\mathrm{ens}}(x^{(k)}), \quad k = 1,\dots,K$, then the stability is quantified by measuring how similar the explanation of the original instance is to the explanations of its perturbed counterparts. 
%We use the Euclidean distance because it is simple and straightforward. 
We then invert the instability to obtain a stability score.

\[
S(x_i) = \frac{1}{K} \sum_{k=1}^{K} \frac{1}{1 + |\phi_{ens}(x_i) - \phi_{ens}(x_i^{(k)})|_2}
\]
A perfectly stable explanation yields $S(x) \approx 1$, while highly inconsistent explanations approach zero.

\subsection{Label-Aware Group Aggregation}\label{group}
To eliminate the risk of misleading disparity estimates, we aggregate group stability using label-aware aggregation. That is, we explicitly condition stability on the true class label before performing any group-level comparisons. Formally, let the complete set of intersectional subgroups be denoted by $\mathcal{G}$,
%as defined in Section \ref{cartessian}, 
and let $y \in \{0,1\}$ denote the binary ground-truth label. 
%For each instance $x$, the stability score is given by $S(x)$ from Eq. \ref{eqns}. 
We first partition the dataset into disjoint label-conditional subsets:
\[
\mathcal{D}_{g,y} = \{ x : A(x) = g, Y(x) = y \},
\]
where $A(x)$ denotes the protected group membership of instance $x$. For each subgroup–label cell $(g,y)$, We compute the average stability:
\[
S(g,y) =
\frac{1}{|\mathcal{D}_{g,y}|}
\sum_{x \in \mathcal{D}_{g,y}} S(x),
\quad
\forall g \in \mathcal{G}, y \in \{0,1\}.
\]
%Label-aware aggregation improves robustness by mitigating distributional imbalances that can occur when a single group has very few examples for a given label, thereby making its stability score estimate noisy. 
We address noisiness that results from such aggregation via empirical-Bayes shrinkage, pulling unstable small-sample estimates toward the label-wise mean:
\[
\tilde{S}(g,y) =
\alpha_{g,y} S(g,y)
+
(1 - \alpha_{g,y}) \bar{S}(y)
\]
where
\begin{equation}
    \alpha_{g,y} = \frac{|\mathcal{D}_{g,y}|}{|\mathcal{D}_{g,y}| + \lambda},
\quad
\bar{S}(y)
\frac{1}{|\mathcal{D}_{y}|}
\sum_{\{x: Y(x)=y)\}} S(x),
\end{equation}
and $(\lambda > 0)$ controls the degree of shrinkage. 
%This formulation ensures that small groups are not unfairly penalized due to high-variance estimates, thereby enabling MESD to scale reliably to settings with many intersectional subgroups.

Finally, we aggregate these label-conditioned scores to produce a single group-level measure:
\begin{equation}
S(g) = \sum_{y \in \{0,1\}} w_y , \tilde{S}(g,y),
\label{eq:grp_meas}
\end{equation}
where
\[
w_y = \frac{|\mathcal{D}_y|}{|\mathcal{D}|}
\]
are label prevalence weights, ensuring that each label influences group-level stability in proportion to its frequency. %This weighted aggregation maintains the semantic meaning of stability across heterogeneous label distributions and prevents artificially inflating or suppressing disparities.

\subsection{CVaR-Weighted Group Disparity (MESD)}
To extract the final MESD metric from the group-level stability $S(g)$, we use Conditional Value-at-Risk (CVaR), which as a robust, risk-sensitive measure that 
gives greater weight to the worst-performing group-level disparities. This reveals subgroups that may exhibit disproportionately low explanation stability. 
Formally, for any pair of intersectional groups $(g_i, g_j \in \mathcal{G})$, define the pairwise stability disparity as
\[
D_{ij} = \left| S(g_i) - S(g_j) \right|,
\]
where $(S(g))$ is the aggregated group-level stability defined in Equation \ref{eq:grp_meas}. To incorporate the severity of instability, we define a risk score for each pair:
\[
R_{ij} = 1 - \min\{ S(g_i), S(g_j) \},
\]
so that pairs involving groups with low procedural fairness receive higher risk values. The collection of risks ${R_{ij}}$ induces a distribution from which MESD computes the $(\alpha)$-tail risk threshold
\[
\tau_\alpha = \mathrm{Quantile}_{1 - \alpha}R_{ij},
\]
identifying the set of ``worst-case'' disparities. CVaR then assigns weights only to disparities whose risk exceeds this threshold:
\[
w_{ij} = \frac{\{\max{ (R_{ij} - \tau_\alpha), 0 \}}}{\sum_{k,\ell} \{\max{( R_{k\ell} - \tau_\alpha), 0 }\} + \varepsilon}.
\]
%This weighting mechanism ensures that MESD does not treat all group pairs equally; instead, it concentrates emphasis on those involving the most vulnerable communities. 
The CVaR-weighted MESD measure is therefore given by:
\begin{equation}\label{mesd_eqn}
\mathrm{MESD}_{\mathrm{CVaR}}(\alpha)
= \sum_{i < j} w_{ij} , D_{ij},
\end{equation}
%yielding an interpretable scalar quantity that 
which summarizes the severity of tail disparities under explanation perturbations. Overall, MESD measures how unevenly explanation stability is distributed across groups, with higher values indicating that some subgroups receive significantly less reliable explanations than others.

MESD can be interpreted as a three-stage robust explanation fairness procedure, where perturbation-based stability captures local explanation consistency, label-aware aggregation aligns procedural fairness with outcome-conditional distributions, and empirical-Bayes shrinkage reduces variance in sparse intersectional cells. 
The CVaR formulation then converts these estimates into a risk-sensitive disparity measure, ensuring that worst-case subgroup instability dominates the metric rather than average behavior.

\begin{algorithm}[t]
\caption{Multi-category Explanation Stability Disparity (MESD)}
\label{alg:mesd}
\begin{algorithmic}[1]
\REQUIRE Model $f$, Data $X$, Labels $Y$, Protected attributes $A_1, \ldots, A_m$, Explainers $E$, Perturbation count $K$, CVaR level $\alpha$, Shrinkage $\lambda$
\ENSURE MESD$_{\text{CVaR}}$ score

\STATE $\mathcal{G} \leftarrow A_1 \times A_2 \times \cdots \times A_m$ \COMMENT{Intersectional subgroups}

\FOR{each instance $x_i \in X$}
    \STATE $\phi_0 \leftarrow \sum_{e \in E} w_e \cdot \phi^{(e)}(x_i)$ \COMMENT{Ensemble attribution}
    \FOR{$k = 1$ to $K$}
        \STATE $x_i^{(k)} \leftarrow \text{Perturb}(x_i)$ \COMMENT{Gaussian + masking}
        \STATE $\phi_k \leftarrow \sum_{e \in E} w_e \cdot \phi^{(e)}(x_i^{(k)})$
    \ENDFOR
    \STATE $S(x_i) = \frac{1}{K} \sum_{k=1}^{K} \frac{1}{1 + |\phi_{ens}(x_i) - \phi_{ens}(x_i^{(k)})|_2}$ \COMMENT{Stability}
\ENDFOR

\FOR{each group $g \in \mathcal{G}$, label $y \in \{0,1\}$}
    \STATE $\bar{S}(g,y) \leftarrow \text{mean}(\{S(x_i) : A(x_i) = g, Y(x_i) = y\})$
    \STATE $\alpha_{g,y} \leftarrow |D_{g,y}| / (|D_{g,y}| + \lambda)$ \COMMENT{EB shrinkage}
    \STATE $\tilde{S}(g,y) \leftarrow \alpha_{g,y} \cdot \bar{S}(g,y) + (1-\alpha_{g,y}) \cdot \bar{S}(y)$
\ENDFOR

\STATE $S(g) \leftarrow \sum_y \frac{|D_y|}{|D|} \cdot \tilde{S}(g,y)$ for each $g$ \COMMENT{Label-aware}

\FOR{each pair $(g_i, g_j) \in \mathcal{G} \times \mathcal{G}, \; i < j$}
    \STATE $D_{ij} \leftarrow |S(g_i) - S(g_j)|$; \quad $R_{ij} \leftarrow 1 - \min(S(g_i), S(g_j))$
\ENDFOR

\STATE $\tau_\alpha \leftarrow \text{Quantile}_{1-\alpha}(\{R_{ij}\})$ \COMMENT{CVaR threshold}
\STATE $w_{ij} \leftarrow \max(R_{ij} - \tau_\alpha, 0) / \sum_{k<\ell} \max(R_{k\ell} - \tau_\alpha, 0)$
\RETURN $\text{MESD}_{\text{CVaR}} = \sum_{i<j} w_{ij} \cdot D_{ij}$
\end{algorithmic}
\end{algorithm}

\section{Optimizing Fairness and Explainability}
To select a fair model with high utility and high explainability, we propose a multi-objective optimization framework, Utility, Explainability, and Fairness (UEF), that optimizes these three objectives using Non-Dominating Sorting Genetic Algorithm (NSGA-II) \cite{deb2002fast}. The goal of our UEF is to achieve a better trade-off between these objectives. In our modeling, each individual in the population encodes a specific model configuration. Algorithm \ref{alg:uef} shows the steps of UEF.

Let $\theta$ denote the decision variables. We define three objectives $F_{\text{perf}}(\theta),  F_{\text{out}}(\theta),  F_{\text{proc}}(\theta)$, which represent performance (utility), outcome-oriented fairness, and procedural-oriented fairness, respectively. 
Each objective is formulated as a quantity to be minimized (the $F_{perf}$ is a maximization problem but was converted to minimization using negation). 
The multi-objective optimization problem is stated as follows:
\begin{equation}
\label{eq:moo}
\min_{\mathbf{\theta} \in {\Theta}} \big( F_{\text{perf}}(\theta),  F_{\text{out}}(\theta),  F_{\text{proc}}(\theta) \big) 
\end{equation}
where $\Theta$ is the space of all valid model configurations. 
%The next section describe the objectives in detail:
%\begin{enumerate}
%\item 

The first objective measures the predictive utility of the model. Let $(f_\theta(x)\in[0,1])$ denote the predicted probability of the positive class. Then
\[
F_{\text{perf}}(\theta) = -\mathrm{AUC}(f_\theta;\mathcal{D}),
\]
where AUC denotes the area under the ROC curve evaluated on a validation set. The negative sign ensures that all objectives are minimized.

%\item 
The second objective captures traditional outcome-oriented fairness, which evaluates disparities in model predictions across protected groups. Let $(a\in\mathcal{A})$ denote a protected attribute (or intersectional group), and let (Y) be the true label. 
We adopt Demographic Parity (DP), which requires that the positive prediction of the model to be invariant across protected groups. For multiple protected categories, demographic parity is defined as:
\[
P(\hat{Y}=1 \mid A=a)=
P(\hat{Y}=1 \mid A=a')
\quad
\forall a,a' \in \mathcal{A}.
\]
In practice, we quantify deviations from this ideal using the maximum pairwise disparity:
\[
F_{\mathrm{out}}(\theta)
\max_{a,a' \in \mathcal{A}}
\big|
P(\hat{Y}=1 \mid A=a)-
P(\hat{Y}=1 \mid A=a')
\big|.
\]
This formulation naturally generalizes binary demographic parity to intersectional settings.

%\item 
The third objective, $F_{proc}(\theta)$, is our MESD procedural metric defined in equation \ref{mesd_eqn}. The aim of MESD is to quantify the difference in explanation disparity between intersectional groups within protected attributes.
%\end{enumerate}

To solve the above multi-objective problem, we used NSGA-II. 
%NSGA-II repeatedly evolves a population of candidate solutions toward a better trade-off among objectives, using the biological principles of genetic variation and natural selection. NSGA-II was selected because it has the capacity to efficiently approximate the Pareto front in a single run. 
In our optimization setup, each population represents a model configuration $\theta$ as defined earlier. We begin by initializing a population $P_0$ of $N$ random solutions, sampling each hyperparameter from its specified range. We evaluate every solution $\theta$ in the population on all the objectives, which means we use the hyperparameter set returned by the objective to train a model, and we evaluate the performance of the model using AUC, DP, and MESD on a test dataset. 

The Pareto front returned by NSGA-II comprises several model configurations.
%, each containing a set of non-dominated solutions that represent different trade-offs among predictive performance, outcome-oriented fairness, and procedural fairness. 
To select a single model for final deployment, we adopt Chebyshev scalarization as a post-optimization decision rule. Chebyshev scalarization minimizes the maximum deviation from an ideal point.

Let $\mathbf{F}(\theta) = (F_\text{perf}(\theta), F_\text{out}(\theta), F_\text{proc}(\theta)) $ denote the vector of objectives 
%corresponding to utility, outcome fairness, and procedural fairness (MESD). 
Let $(\mathbf{z}^\star = (z_\text{perf}^\star, z_\text{out}^\star, z_\text{proc}^\star))$ be an ideal point. Chebyshev scalarization selects the solution
\[
\theta^\star =
\underset{\theta \in \mathcal{P}}{\text{argmin}}
\max_{k \in \lbrace \text{perf,out,proc}\rbrace}
\lambda_k , \lvert F_k(\theta) - z_k^\star \rvert ,
\]
where $(\lambda_k > 0)$ are optional scaling coefficients used to normalize objectives or encode relative importance.
%, this formulation prioritizes balanced solutions by penalizing models that perform extremely poorly on any single objective, even if they excel on others.

\begin{algorithm}[t]
\caption{UEF: Utility-Explanation-Fairness Optimization}
\label{alg:uef}
\begin{algorithmic}[1]
\REQUIRE Training data $(X_{\text{tr}}, Y_{\text{tr}})$, Test data $(X_{\text{te}}, Y_{\text{te}})$, Sensitive attrs $A$, Pop size $N$, Generations $T$, Chebyshev weights $\boldsymbol{\lambda}$
\ENSURE Selected model $f^*$

\STATE Initialize population $P_0$ of $N$ random configurations $\theta_i$
\STATE Each $\theta_i = (\text{lr}, \ell_2, \text{dropout}, \text{epochs}, \text{threshold})$

\FOR{generation $t = 1$ to $T$}
    \FOR{each $\theta \in P_{t-1}$}
        \STATE Train model $f_\theta$ on $(X_{\text{tr}}, Y_{\text{tr}})$ using $\theta$
        \STATE $F_1(\theta) \leftarrow -\text{AUC}(f_\theta, X_{\text{te}}, Y_{\text{te}})$ \COMMENT{Utility}
        \STATE $F_2(\theta) \leftarrow \text{DP}_{\text{int}}(f_\theta, A)$ \COMMENT{Outcome fairness}
        \STATE $F_3(\theta) \leftarrow \text{MESD}_{\text{CVaR}}(f_\theta)$ \COMMENT{Alg.~\ref{alg:mesd}}
    \ENDFOR
    \STATE $P_t \leftarrow \text{NSGA-II}(P_{t-1}, \{F_1, F_2, F_3\})$ \COMMENT{Non-dominated sort}
\ENDFOR

\STATE $\mathcal{P} \leftarrow$ Pareto front from $P_T$
\STATE $z^* \leftarrow (\min F_1, \min F_2, \min F_3)$ \COMMENT{Ideal point}
\STATE $z^{\text{nad}} \leftarrow (\max F_1, \max F_2, \max F_3)$ \COMMENT{Nadir point}

\STATE $\theta^* \leftarrow \arg\min_{\theta \in \mathcal{P}} \max_{k} \lambda_k \cdot \frac{|F_k(\theta) - z^*_k|}{z^{\text{nad}}_k - z^*_k}$ \COMMENT{Chebyshev}

\RETURN $f^* \leftarrow$ retrain and evaluate $\theta^*$ with full MESD
\end{algorithmic}
\end{algorithm}

\section{Experimental Evaluation}
We evaluate MESD and UEF through seven experiments, each addressing a specific research question. We use three benchmark datasets from UCI repository \cite{asuncion2007uci}: German Credit (2 protected attributes: age, sex), Recidivism (2: race, gender), and Adult Income (3: age, race, sex).

\subsection{Baselines}
We compare our method (UEF) with four methods, which are ERM (vanilla neural network without fairness intervention), Reweighing (RW) \cite{jiang2020identifying}, Reductions (RD) \cite{agarwal2018reductions}, and Adversarial Debiasing (ADL) \cite{zhang2018mitigating}, all having fairness intervention. To control for architecture, all methods use an identical neural network, and we tuned their parameters using random search. All experiments run across 5-fold cross-validation, reporting mean $\pm$ standard deviation.

\subsection{Evaluation Metrics}
AUC, F1, Demographic Parity gap (DP), Equalized Odds gap (EOD), MESD

\subsection{Research Questions}
We report several experiments to answer the following questions:
\begin{enumerate}
    \item [RQ1] Does UEF effectively balance utility, outcome fairness, and procedural fairness compared to baselines?
    \item [RQ2] How does UEF perform in terms of Pareto efficiency (non-dominance) across datasets compared to baselines?
    \item [RQ3] How does CVaR-weighted MESD improve detection of worst-case subgroup disparities?
    \item [RQ4] How sensitive is MESD to hyperparameters (perturbation, explainers, baseline)?
\end{enumerate}

\subsection{Experimental Results}

\begin{table*}[t]
\centering
\caption{Comparative evaluation across three datasets using 5-fold stratified cross-validation (mean $\pm$ std). 
$\uparrow$ higher is better, $\downarrow$ lower is better. \textbf{Bold} indicates best performance per metric per dataset.}
\label{tab:rq1}
\small
\begin{tabular}{ll ccccc}
\toprule
\textbf{Dataset} & \textbf{Method} & \textbf{AUC} $\uparrow$ & \textbf{F1} $\uparrow$ & \textbf{DP} $\downarrow$ & \textbf{EOD} $\downarrow$ & \textbf{MESD} $\downarrow$ \\
\midrule
\multirow{5}{*}{Adult Income} & ERM & {0.900$\pm$0.004} & {0.652$\pm$0.018} & 0.305$\pm$0.021 & 0.772$\pm$0.210 & 0.0012$\pm$0.0002 \\
 & RW & 0.886$\pm$0.004 & 0.573$\pm$0.024 & 0.165$\pm$0.020 & 0.715$\pm$0.183 & 0.0013$\pm$0.0001 \\
 & RD & 0.669$\pm$0.009 & 0.501$\pm$0.017 & \textbf{0.064$\pm$0.007} & 0.735$\pm$0.150 & 0.0043$\pm$0.0017 \\
 & ADL & 0.889$\pm$0.009 & 0.628$\pm$0.016 & 0.245$\pm$0.013 & 0.637$\pm$0.100 & 0.0016$\pm$0.0005 \\
 & UEF & \textbf{0.906$\pm$0.002} & \textbf{0.722$\pm$0.037} & 0.127$\pm$0.069 & \textbf{0.526$\pm$0.294} & \textbf{0.0006$\pm$0.0005} \\
\midrule
\multirow{5}{*}{Recidivism} & ERM & 0.998$\pm$0.001 & \textbf{0.988$\pm$0.002} & 0.260$\pm$0.042 & 0.042$\pm$0.040 & 0.0156$\pm$0.0293 \\
 & RW & 0.998$\pm$0.000 & 0.987$\pm$0.001 & 0.257$\pm$0.045 & \textbf{0.032$\pm$0.035} & 0.0153$\pm$0.0288 \\
 & RD & 0.899$\pm$0.005 & 0.879$\pm$0.005 & {0.073$\pm$0.033} & 0.350$\pm$0.116 & {0.0163$\pm$0.0088} \\
 & ADL & \textbf{0.999$\pm$0.001} & 0.984$\pm$0.001 & 0.257$\pm$0.045 & 0.038$\pm$0.038 & 0.0146$\pm$0.0277 \\
 & UEF & 0.998$\pm$0.017 & 0.937$\pm$0.095 & \textbf{0.055$\pm$0.017} & 0.097$\pm$0.094 & \textbf{0.0034$\pm$0.0030} \\
\midrule
\multirow{5}{*}{German Credit} & ERM & 0.768$\pm$0.034 & 0.519$\pm$0.041 & 0.315$\pm$0.100 & 0.416$\pm$0.133 & 0.0006$\pm$0.0002 \\
 & RW & 0.763$\pm$0.035 & 0.497$\pm$0.021 & 0.262$\pm$0.088 & 0.470$\pm$0.190 & 0.0007$\pm$0.0002 \\
 & RD & 0.658$\pm$0.039 & 0.512$\pm$0.061 & \textbf{0.146$\pm$0.049} & 0.355$\pm$0.039 & 0.0015$\pm$0.0011 \\
 & ADL & \textbf{0.772$\pm$0.042} & {0.521$\pm$0.036} & 0.201$\pm$0.104 & \textbf{0.288$\pm$0.098} & 0.0005$\pm$0.0001 \\
 & UEF & 0.765$\pm$0.034 & \textbf{0.613$\pm$0.128} & 0.187$\pm$0.069 & 0.376$\pm$0.112 & \textbf{0.0004$\pm$0.0002} \\
\bottomrule
\end{tabular}
\end{table*}

Table~\ref{tab:rq1} reports the experimental results across three datasets using 5-fold stratified cross-validation. The results show a consistent pattern, which is that methods that optimize for a single objective excel on that dimension but degrade on others, while UEF achieves the most balanced performance across all three objectives. On Adult Income, the most challenging dataset with three protected attributes, UEF achieves the highest AUC, F1, and the lowest EOD and MESD, demonstrating that multi-objective optimization can simultaneously improve predictive performance, outcome, and procedural fairness, a finding consistent with recent theoretical work on the fairness--accuracy Pareto front \cite{wei2022fairness}. %ERM achieves competitive utility but exhibits the worst outcome fairness, while RD achieves the best DP (0.064) at a severe utility cost (AUC = 0.669). Notably, UEF also produces the best EOD (0.526), suggesting that jointly optimizing for procedural fairness acts as a regularizer that incidentally improves outcome fairness, 

On Recidivism, where all methods achieve near-perfect AUC ($>0.99)$, the differentiation shifts to fairness metrics. UEF again achieves the best DP and MESD, which shows that UEF benefits from the multi-objective of both outcome and procedural fairness. However, methods like RD that was optimize only for outcome fairness achieve the lowest EOD but high MESD value, which demonstrates that it is possible for a method to perform very well on outcome fairness but poorly on procedural fairness, which is motivation for this research.

On German Credit, the smallest dataset, UEF achieves the best MESD and F1. Across all three datasets, no single baseline dominates on all metrics simultaneously. Methods like RD and ADL excel in one dimension of the metric while losing out on the other. Also, no single method outperforms UEF on all MESD, showing that UEF uses more consistent reasoning compared to other methods.

\subsection{Trade-off Between Objectives}
\begin{figure*}
    \centering
    \includegraphics[width=1.0\linewidth]{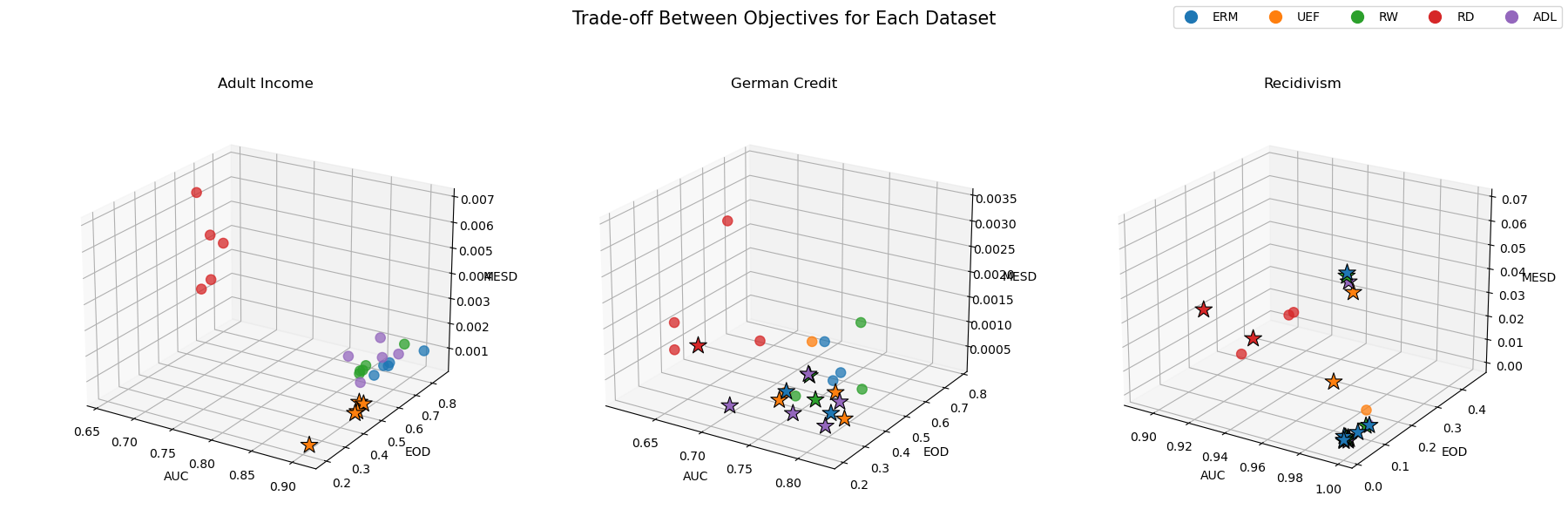}
    \caption{Performance of each algorithm on each dataset}
    \label{fig:pareto}
\end{figure*}

To evaluate each method's ability to produce balanced solutions across all three objectives simultaneously, we assess Pareto non-dominance. A solution is non-dominated on a given fold if no other method achieves simultaneously higher AUC, lower EOD, and lower MESD on the same data split. Table~\ref{tab:pareto} reports the results across 5 folds and 3 datasets (15 evaluations total), and Figure~\ref{fig:pareto} visualizes the three-dimensional objective space. UEF produces 12 out of 15 non-dominated solutions. On the Adult Income dataset, the most challenging setting with three protected attributes yielding eight intersectional subgroups, UEF is the \emph{only} method that achieves non-dominated solutions on all the 5 folds. This result is particularly significant because Adult Income is the dataset where intersectional fairness gerrymandering is most likely to occur, and it demonstrates that multi-objective optimization with MESD is most valuable precisely where the fairness problem is hardest. 
On the German Credit dataset, UEF achieves 3 out of 5 non-dominated solutions, ranking second behind ADL (5), which benefits from the smaller intersectional space (4 subgroups) where adversarial debiasing is particularly effective. On Recidivism, ERM and RW lead with 5 non-dominated solutions each, benefiting from the dataset's near-perfect separability, which reduces the optimization problem to fairness dimensions where simpler methods suffice. However, UEF still achieves 4 non-dominated solutions in this setting. 

Overall, these results demonstrate that UEF's advantage grows with the complexity of the intersectional fairness problem, and UEF fully dominates all baselines, validating the motivation that jointly optimizing for procedural fairness is most impactful where single-axis approaches are most likely to fail.

\begin{table}[t]
\centering
\caption{Number of non-dominated solutions (out of 5 folds) per method and dataset.}
\label{tab:pareto}
\small
\begin{tabular}{l ccc c}
\toprule
\textbf{Method} & \textbf{Recidivism} & \textbf{German} & \textbf{Adult} & \textbf{Total (/15)} \\
\midrule
\textbf{UEF} & 4 & 3 & \textbf{5} & \textbf{12} \\
ADL & 3 & \textbf{5} & 0 & 8 \\
ERM & \textbf{5} & 2 & 0 & 7 \\
RW & \textbf{5} & 2 & 0 & 7 \\
RD & 2 & 1 & 0 & 3 \\
\bottomrule
\end{tabular}
\end{table}

\section{Ablation Study}
\subsection{Effect of CVaR on MESD}
To calculate the final MESD metric, several approaches can be used, such as calculating the variance of each group's stability, the difference between the smallest and largest group stabilities, or the CVaR-weighted MESD. We used CVaR-weighted adjustment to align MESD sensitivity with the worst group, so that MESD would not seem fair on the surface, but one or more groups still suffer from extreme procedural bias. 

\begin{figure}
    \centering
    \includegraphics[width=0.99\linewidth]{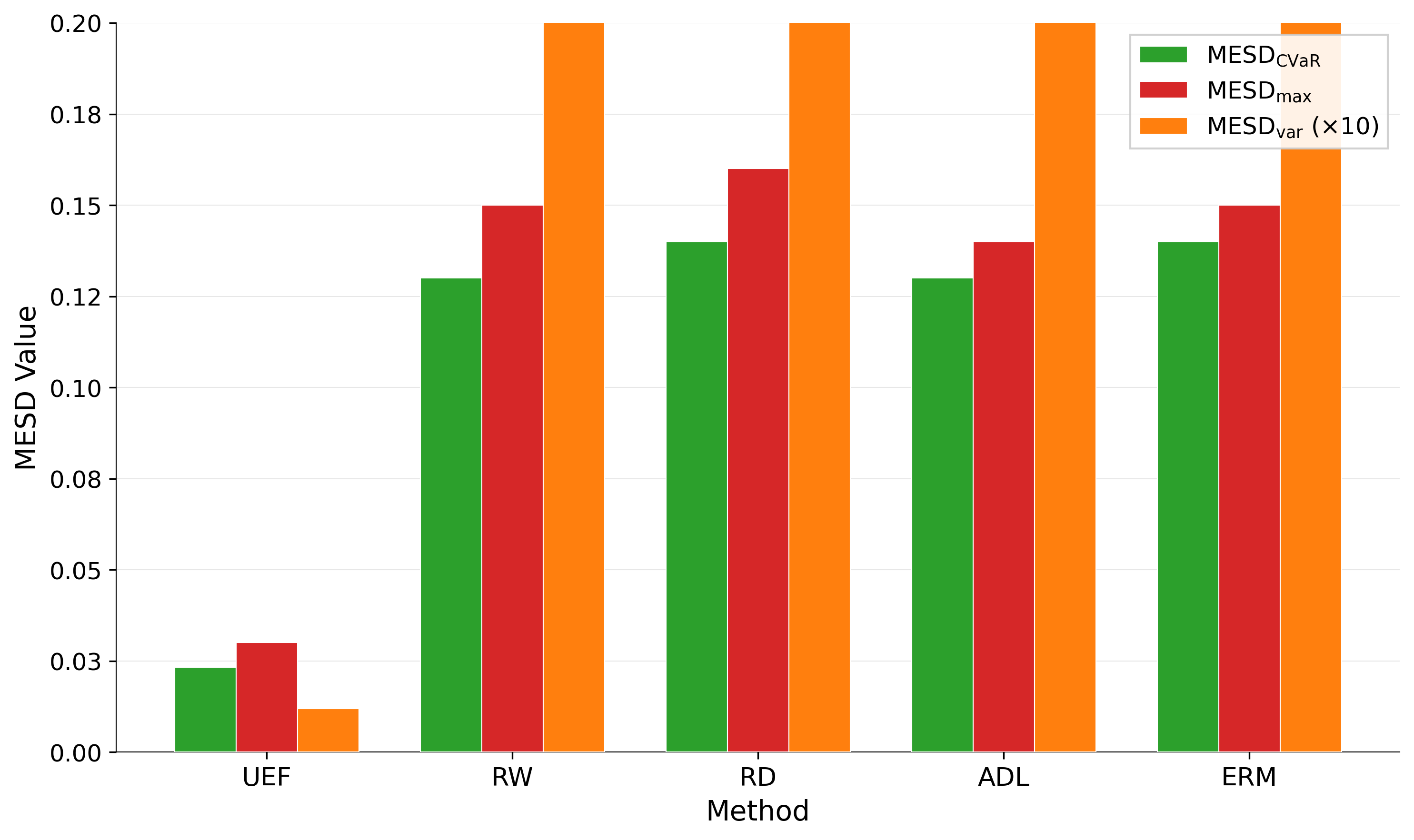}
    \caption{MESD variants from each algorithm on the Recidivism Dataset}
    \label{fig:mesd_variant}
\end{figure}

To test the effect of CVaR, we first show the stability of each intersectional group across all algorithms on the Recidivism dataset in Table \ref{grp}. The Recidivism dataset contains two protected categories: Race (white and other) and Gender (male and female). The Cartesian product of these protected groups yielded four intersectional subgroups. From the group stability results in Table \ref{grp}, we see that one of the groups exhibits reduced stability, which can skew the final MESD metric if the CVaR-weighted method is not used. Secondly, we generated three variants of MESD, which were $MESD\_CVaR$ (which is the original MESD used in Equation \ref{mesd_eqn}), $MESD\_max$ (the maximum difference between any two group stabilities), and $MESD\_var$ (the variance between the differences between the pairwise group stabilities). We plot the results in Figure \ref{fig:mesd_variant}, which allows us to make two points. First, using $MESD\_var$ could yield misleading or overly fair results, masking the worst-case disparity shown in Table \ref{grp}. Second, UEF generates the lowest MESD, resulting from the multi-objective optimization framework

\begin{table}[t]
\centering
\caption{Stability of each algorithm on the Recidivism dataset}
\label{grp}
\begin{tabular}{l ccc cc}
\toprule
Group & UEF & RW & RD & ADL & ERM \\
\midrule
1,0 & 0.89 & 0.74 & 0.73 & 0.74 & 0.73 \\
1,1   & 0.91 & 0.85 & 0.88 & 0.86 & 0.85\\
0,0   & 0.91 & 0.87 & 0.84 & 0.87 & 0.88\\
0,1 & 0.92 & 0.89 & 0.89 & 0.88  & 0.88\\
\bottomrule
\end{tabular}
\end{table}

\subsection{Effect of Empirical-Bayes Shrinkage on MESD}
\begin{figure}[t]
    \centering
    \includegraphics[width=0.99\linewidth]{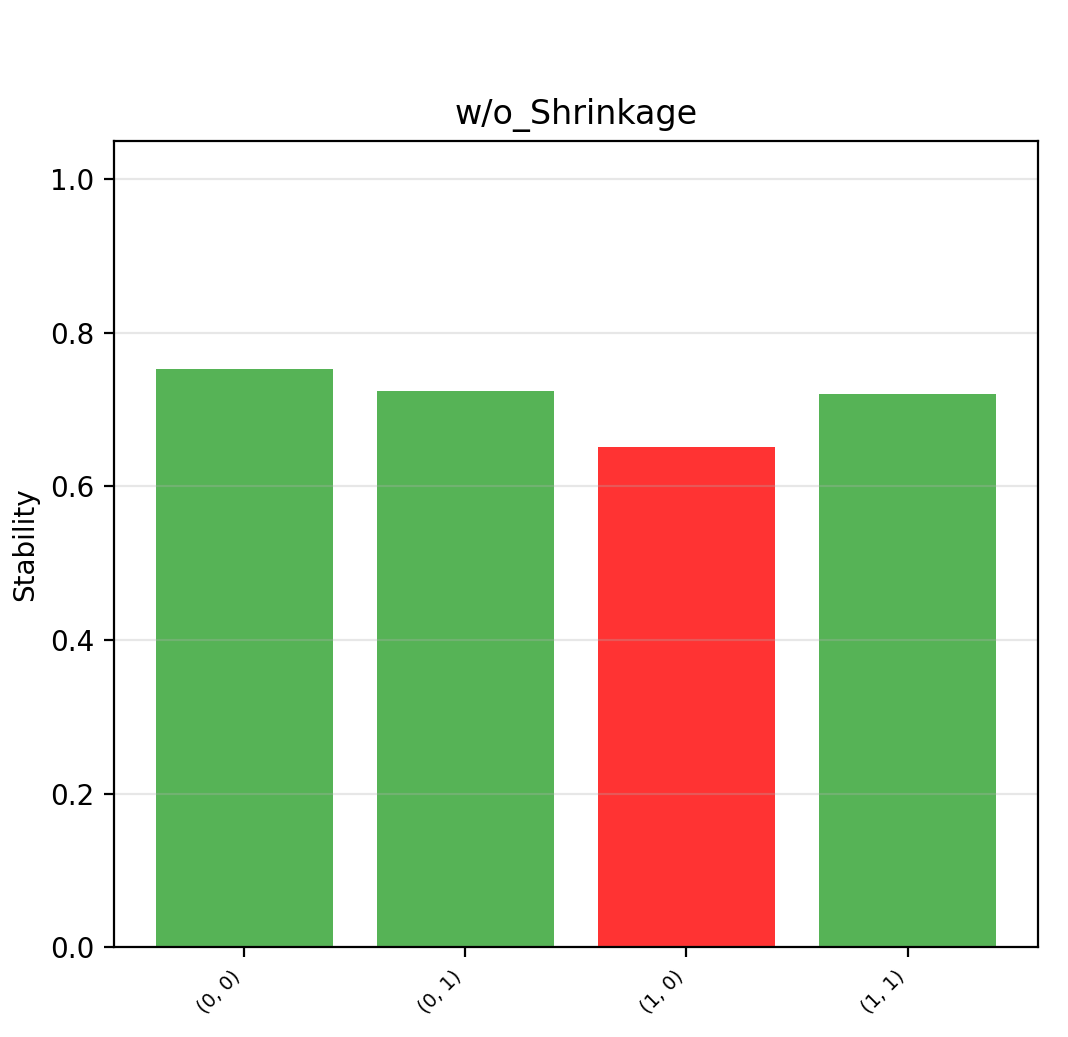}
    \caption{Effect of Empirical-Bayes Shrinkage on Stability of Recidivism}
    \label{fig:mesd_shrinkage}
\end{figure}

Figure~\ref{fig:mesd_shrinkage} isolates the effect of empirical-Bayes (EB) shrinkage on per-group stability estimates for the Recidivism dataset. Without shrinkage, group (1,0) exhibits a stability of 0.65, substantially lower than the other three groups (0.72--0.75). This 0.10-point gap arises as a result of the group having a small number of samples, which can lead to high instability. When EB shrinkage is applied (Full MESD), the estimate for group (1,\,0) is regularized toward the global label-wise mean, producing a more stable and reliable estimate. This result empirically validates the design choice of incorporating shrinkage into MESD.

%\subsection{Baseline Perturbation}
%Figure~\ref{fig:baseline} evaluates MESD's sensitivity to the choice of perturbation baseline value. We test three baselines: zero, feature mean, and feature median, on the Recidivism dataset (results are consistent across all three datasets). The zero and mean baselines produce nearly identical MESD scores, which is expected since the data is standardized with zero mean prior to MESD computation, making the zero baseline and mean baseline functionally equivalent. The median baseline yields slightly lower values, reflecting the fact that median imputation introduces smaller distributional shifts than mean or zero imputation in the presence of skewed features. 

\subsection{MESD Sensitivity to Hyperparameter Choices}
\begin{figure*}[h]
    \centering
    \includegraphics[width=0.99\linewidth]{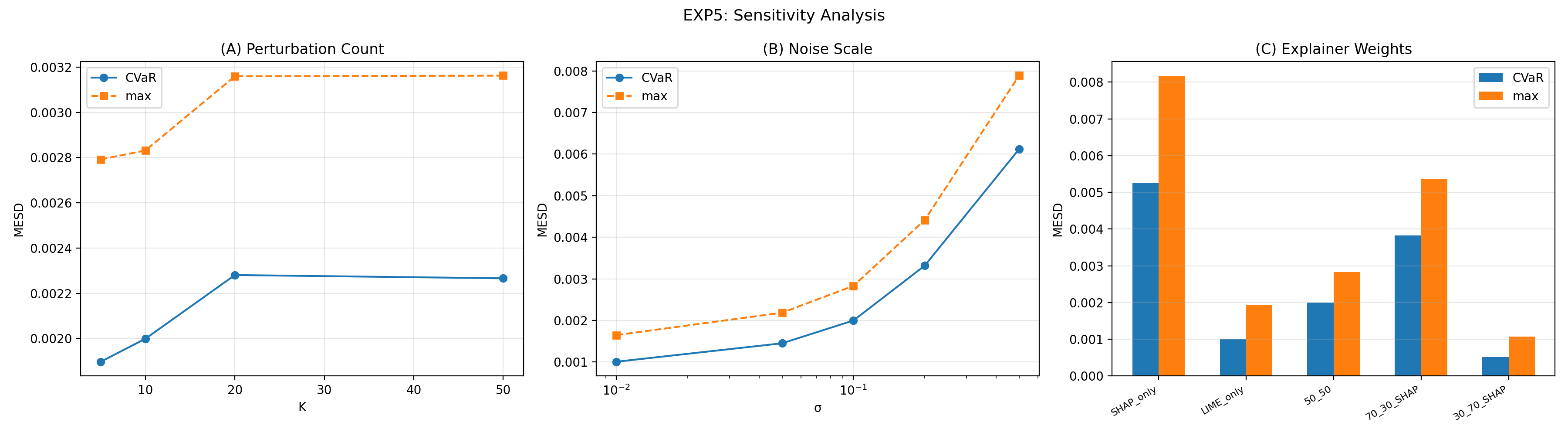}
    \caption{MESD sensitivity on Hyperparameter Choices on Adult Income Dataset}
    \label{fig:sens}
\end{figure*}

Figure~\ref{fig:sens} evaluates the robustness of MESD to three key hyperparameter choices on the Adult Income dataset. Panel~(A) shows that MESD$_{\text{CVaR}}$ increases from 0.0019 at $K=5$ to 0.0023 at $K=20$, then stabilizes between $K=20$ and $K=50$, indicating that too few perturbations underestimates explanation instability due to sampling variance.
Panel (B) reveals that both MESDs increase monotonically with Gaussian noise scale $\sigma$, but at markedly different rates: MESD$_{\max}$ grows from 0.002 at $\sigma=0.01$ to 0.008 at $\sigma=0.5$, while MESD$_{\text{CVaR}}$ grows more gradually from 0.001 to 0.006, demonstrating that CVaR-weighted aggregation is more resistant to noise amplification than the max-gap variant. Also, confirms that perturbations should remain within a small radius around the original input because large $\sigma$ pushes perturbed instances into distant regions of the feature space where explanation divergence reflects genuine distribution shift rather than procedural instability. 
 
Panel (C) reveals a pronounced asymmetry in explainer behavior. SHAP-only produces the highest MESD values, indicating that SHAP attributions are inherently more sensitive to perturbation than LIME attributions. The 50/50 ensemble produces intermediate values that are lower than the arithmetic mean of the two single-explainer scores, confirming that ensembling has a stabilizing effect beyond simple averaging, consistent with findings by Bhatt et al.~\cite{bhatt2020evaluating} that explanation aggregation reduces adversarial vulnerability. 

\begin{figure}[t]
    \centering
    \includegraphics[width=0.99\linewidth]{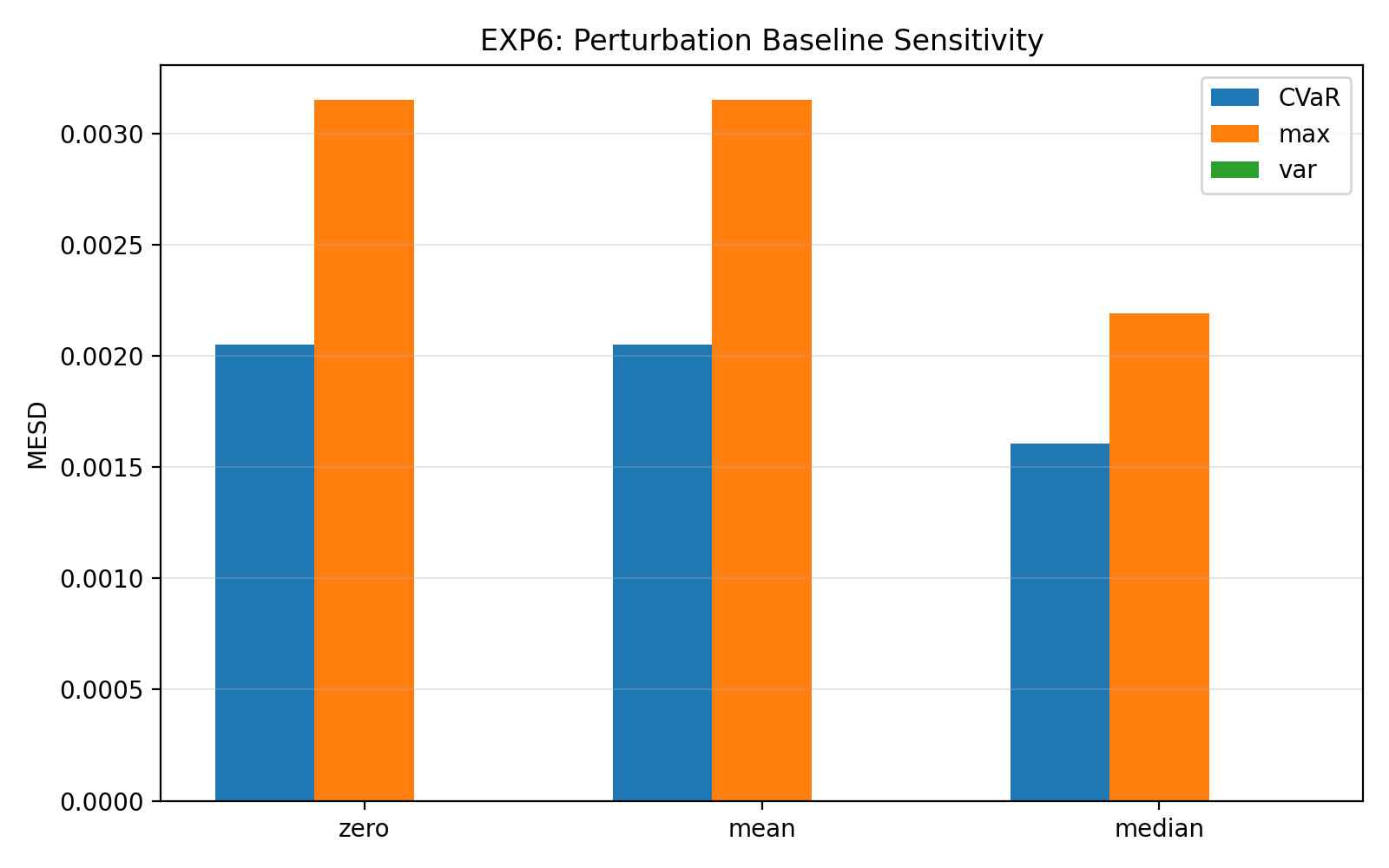}
    \caption{Baseline Perturbation}
    \label{fig:baseline}
\end{figure}

Figure \ref{fig:baseline} shows that MESD is invariant to the choice of perturbation baseline (zero, mean, and median).

\subsection{Effect of Chebyshev Weights on  Objectives}

\begin{figure*}[t]
    \centering
    \includegraphics[width=0.99\linewidth]{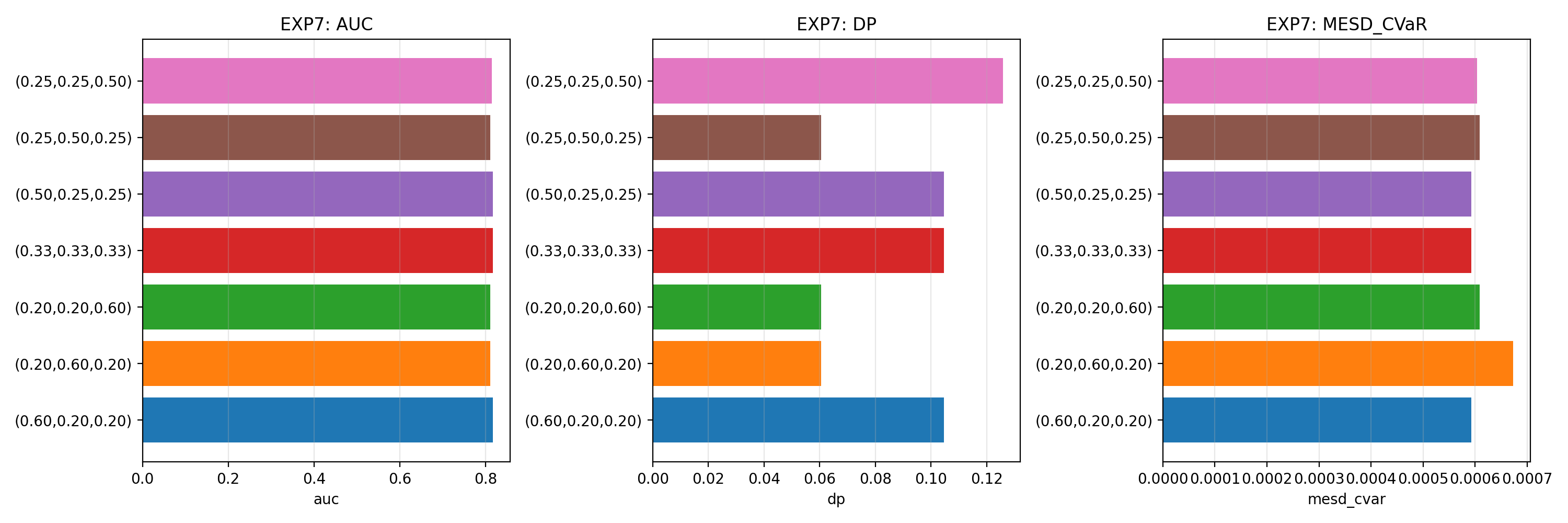}
    \caption{Chebyshev Scalarization}
    \label{fig:chev}
\end{figure*}

Figure~\ref{fig:chev} evaluates how varying the Chebyshev scalarization weights $(\lambda_{\text{perf}}, \lambda_{\text{fair}}, \lambda_{\text{proc}})$ affects the model selected from the Pareto front, shown here on the German Credit dataset (results are consistent across all datasets).
%The weight triplets represent the relative importance assigned to utility (AUC), outcome fairness (DP), and procedural fairness (MESD) during post-optimization selection. 
Three key observations emerge. First, AUC remains remarkably stable across all weight configurations, indicating that the Pareto front produced by NSGA-II contains solutions with consistently strong predictive performance. Second, the trade-off between DP and MESD is clearly visible the utility-heavy configuration (0.60,\,0.20,\,0.20) yields the highest DP ($\sim$0.11) and lowest MESD ($\sim$0.0001), while the fairness-heavy configuration (0.25,\,0.25,\,0.50) achieves the lowest DP ($\sim$0.02) but the highest MESD ($\sim$0.0007). This confirms that the Pareto front captures a genuine three-way trade-off rather than collapsing to a single dominant solution. Third, the equal-weight configuration (0.33,\,0.33,\,0.33), which we adopt as the default throughout our experiments, selects a solution in the middle of both the DP and MESD ranges.

\section{Discussion}
Our results demonstrate that outcome-oriented fairness and procedural fairness are distinct dimensions that can diverge in practice. Specifically, our experiments in Table \ref{tab:rq1} show that methods that were specifically optimized for outcome-oriented fairness are unable to mitigate procedural fairness. Hence, they are not able to generate a holistic model that is fair in all three dimensions. Also, models that use systematically different reasoning for different groups create divergent incentive structures. For instance, if a credit model rewards income heavily for one intersectional group but penalizes debt for another, members of those groups receive fundamentally different signals about what behaviors lead to approval, perpetuating the inequalities the outcome-oriented fairness intervention was designed to correct.

The Pareto dominance analysis provides empirical evidence for treating fairness as a multi-objective problem. UEF produces non-dominated solutions on 12 out of 15 folds, and it is the only method that achieves a non-dominated solution on the Adult Income dataset, the setting with eight intersectional subgroups where the optimization landscape is most complex. This suggests that the benefit of multi-objective optimization scales with the difficulty of the intersectional fairness problem, especially where single-objective approaches are most likely to miss procedurally unfair solutions that appear distributionally acceptable. The ablation study shows that without empirical-Bayes shrinkage, group (1, 0) on Recidivism with the smallest samples drops to a stability of 0.65 while other groups remain at 0.72--0.75, a disparity that the full MESD metric detects and that outcome metrics structurally cannot
The sensitivity analyses further establish MESD as a practical auditing tool, and they show that the metric is stable across perturbation counts, robust to the choice of perturbation baseline, and tunable via explainer weights to accommodate application-specific requirements. These results position MESD as a necessary component of a procedural fairness audit. However, MESD should be deployed alongside domain expertise, participatory oversight, and complementary interpretability methods as AI systems are increasingly subject to regulatory scrutiny under frameworks such as the EU AI Act. Also, the results support the procedural justice argument advanced by \cite{thibaut1975procedural} that fair outcomes arrived at through inconsistent reasoning processes undermine the legitimacy of the decision system, particularly for individuals at the intersection of multiple marginalized identities who are most vulnerable to fairness gerrymandering \cite{kearns2018preventing}.

\section{Conclusion}
We introduced Multi-category Explanation Stability Disparity (MESD), a procedural fairness metric that detects disparities in explanation quality across intersectional subgroups formed by the Cartesian product of multiple protected attributes. MESD integrates three components, which are label-aware aggregation, empirical-Bayes shrinkage to stabilize estimates for small intersectional groups, and CVaR-weighted tail-risk measurement to prevent fairness gerrymandering. By embedding MESD within the UEF multi-objective optimization framework, we showed that jointly optimizing utility, outcome fairness, and procedural fairness yields models that are non-dominated across all three objectives more consistently than any baseline. UEF produced non-dominated solutions on 12 out of 15 evaluations across three datasets, and was the only method to achieve non-domination on the most intersectionally complex dataset.
%Our sensitivity analyses confirmed that MESD is robust to several hyperparameter choices

%sveral perturbation hyperparameters, baseline choice, and explainer weighting, establishing it as a practical tool for procedural fairness auditing. 

\section{Limitations and Future Work}
This work has several limitations that suggest directions for future research. First, MESD relies on post-hoc explanation methods (SHAP and LIME), which are imperfect approximations of a model's internal reasoning \cite{slack2020fooling}. A model producing stable post-hoc explanations is not guaranteed to be reasoning consistently at a mechanistic level. Future work should explore in-processing explainability methods, such as layer-wise relevance propagation or attention-based attributions, that more directly reflect the model's decision process. Second, our evaluation is limited to tabular datasets with binary protected attributes (two races), and extending MESD to multi-level attributes, continuous protected features, and non-tabular modalities such as text and images remains an important open problem. Third, the computational cost of MESD within the NSGA-II optimization can be reduced by using more efficient explanation methods such as FastSHAP~\cite{jethani2021fastshap} or amortized explainers. 
%Finally, as with all technical fairness interventions~\cite{green2020false, binns2018fairness}, MESD should be deployed as one component within a broader accountability framework.
%that includes participatory design with affected communities, domain-specific regulatory guidance, and ongoing human oversight rather than as a standalone solution to procedural discrimination.
\bibliography{aaai25.bib}

\end{document}